# Neurocognitive Informatics Manifesto

Włodzisław Duch
Department of Informatics, Nicolaus Copernicus University,
Grudziądzka 5, Toruń, Poland.
Google: Włodzisław Duch

**Abstract**
Informatics studies all aspects of the structure of natural and artificial information systems. Theoretical and abstract approaches to information have made great advances, but human information processing is still unmatched in many areas, including information management, representation and understanding. Neurocognitive informatics is a new, emerging field that should help to improve the matching of artificial and natural systems, and inspire better computational algorithms to solve problems that are still beyond the reach of machines. In this position paper examples of neurocognitive inspirations and promising directions in this area are given.

## 1. Introduction

Technology is developing at an unprecedented pace, from simple systems that were easy to control to handheld devices, such as personal digital assistant (PDA) computers or smart telephones with multimedia capabilities, devices with so many options that rarely more than a tiny fraction are ever used even by quite intelligent people. Further miniaturization of integrated circuits will only increase complexity of the environment, exceeding capability of cognitive information processing by most humans. In other areas situation is not better: only a small fraction of all functions is used in a typical word processor or other software, equipment used for medical diagnosis is capable of many sophisticated measurements, only a fraction of which is understood and relied on by medical doctors. Human brain is highly adaptable due to the neuroplasticity [1], therefore the speed of reaction and the ability to deal with complex devices is increasing, but very few people can become experts in more than one area.

The increasing dependence on the sophisticated technology that should help with the complexities of modern life poses requires revolutionary changes in organization of information. The great challenge for informatics is thus in creation of interfaces that will make technology more human, easier to use, adjusted to our ways of information processing. Humanized interfaces imply the ability of software to anticipate human reactions, understand questions and errors that humans are prone to make, ask questions to specify precisely the request, in general adjust itself to the human-like ways of thinking. These problems are only partially addressed by the current branches of informatics, although there have been some attempts in the past to understand the ways our brains are processing information and to adjust are methods accordingly. Already H. Spencer in "Principles of Psychology'" published in 1855 made a conjecture that all intelligence may be interpreted in terms of successful associations between psychological states driven by the strength of connections between the internal states (see the early history of connectionism in [2]). Yet to this day we do not have machines that would help us to find interesting associations, coupling to and enhancing our natural abilities. The methods that are currently used to represent knowledge [3][4] are not as flexible as natural language and the imagery that brains are capable of. Only human brains are capable of using language, and bringing the existing Natural Language Processing methods closer to what human brains do reacting to and understanding concepts expressed by language structures is ultimately the only certain way to human-level competence in this area. While formal and statistical models of language may capture some important features it is unlikely that the texts itself contain sufficient information to train a disembodied system to fully understand the meaning.

There is general agreement among philosophers of language and neuroscientists who work on language [5] that the meaning of basic concepts is grounded in the embodiment, the ability to perceive and act. The basic function of the language in the brain is to activate perception/action networks. This trend started in robotics with R. Brooks Cog project manifesto [6][7], in philosophy with F. Varela, E. Thompson and E. Rosch 1991 book [8], and in linguistics it has been represented by G. Lakoff and others [9][10]. Problems with traditional approach to natural language led some cognitive systems experts to base their hope on developmental approach, hoping that robots acting in the world are going to develop some proto-language exchanging messages [11][12]. Such efforts are not new, already in 1998 Hutchens and Alder created a chatterbot called MegaHAL [13] based on stochastic language model that could be trained by interaction with people, but after 10 years this chatterbot is still on the level of using 2-3 word combinations that can be interpreted as capturing some statistical relations from the conversations with humans. There has been some interesting progress in modeling relations between word use, perceptions and actions, relating words to their real referents and helping to define word meanings in specific contexts. These models are still very much restricted to a few areas, such as spatial relations, or color and shape naming, but can be used as an inspiration for simpler representations. Perceptual properties of objects are associated with possible actions that involve these objects, including motor affordances for manipulation. A ball has a round shape, color, size, but also may be kicked or bounced. All this information should be included in the representation of the "ball" concept.

Developmental approach, as well as fully embodied approaches may provide useful inspirations but are not directly applicable to large scale practical NLP applications. The challenge is to identify the processes and find approximations to what brains are doing and use them to correct and enhance the existing vector-based approaches. This is the foundation for more detailed understanding of higher cognitive processes related to thinking and problem solving. As Feldman writes [14] "thought is structured neural activity", and "language is inextricable from thought and experience". Formal reasoning and linguistic methods are quite useful but quite different from information processing that our brains do.

The goal of neurocognitive informatics is to draw inspirations from recent advances in neuroscience, create practical algorithms, especially in areas where traditional approaches fail, and increase our understanding of higher cognitive processes by providing simple models that explain how our minds work. In "The Cognitive Neurosciences" book [15] that largely defined the whole field of research, Gazzaniga wrote: "The future of the field, however, is in working toward a science that truly relates brain and cognition in a mechanistic way." Various approaches to do that will be considered here. In the next chapter relation of neurocognitive informatics to other, existing fields is outlined. The third chapter is mainly focused on language and associative processes. In chapter four some results inspired by the neurocognitive approach to language are presented. The final chapter presents some conclusions and visions of the future of this field.

## 2. Neurocognitive informatics and related fields

Inspiration from nature have of course always been used in science, including computer science. Computational intelligence emerged from interactions of several research communities with overlapping inter-ests, inspired by observations of natural information processing [16]. The three main branches of computational intelligence include evolutionary computing, neural networks and fuzzy logic. Evolutionary processes that structured sensory organs, brains and intelligence are used as an inspiration for various genetic optimization algorithms [17][18]. Later also other biologically-inspired optimization approaches, such as ant, swarm, and immunological system algorithms, have been introduced [19]-[21]. All these biologically inspired algorithms are used for similar applications, although the time-scales of evolutionary, behavioral and immunological processes are very different, and the type of intelligence they are related to is also quite different.

The field of neural networks is quite broad, but mostly inspired by models of single neurons that perform basic computations, models of perception and sensorimotor reactions (primary sensory and motor cortices), a threshold-based perceptron, or spiking neuron being the main computational element. Artificial neural networks (ANNs) are usually based on drastically simplified models of biophysical neurons. Initially logical or graded response (sigmoidal) neurons were almost exclusively used in ANN models [22][23] applied for classification, approximation or associative memories. Multi-layer perceptrons (MLP networks) are still the most successful ANN type of models. Vector quantization approaches used for clusterization and self-organization have also been included [24] to account for unsupervised learning and model development of topographical maps. These model use prototypes and similarity functions that could be weighted in a non-linear way, usually using Gaussian factors, to account for non-linear nature of neural interactions. This led to introduction of basis set expansion techniques, used in approximation theory [25] and quite common in pattern recognition, with computation flow that may be presented in a graphical form similar to the MLP networks, and thus became "a neural network" [23], with radial basis function (RBF) networks [26] becoming a major alternative to multilayer perceptron networks (MLPs). Although ANNs in principle could approximate any function there are some relatively simple problems that such networks are not able to solve, requiring exponentially large number of examples [27], for example the topological invariants of patterns, the problem of connectedness (determining if the pattern is connected or disconnected). Such problems can only be solved with a different type of neurons that include at least one phase-sensitive parameter [28], or with spiking neurons [29]. Although feedforward artificial neural networks found wide applications in knowledge extraction from data [30], simulating human perception and recognition of objects and rules, it has not been used much for knowledge representation or reasoning based on such perceptions.

Computational cognitive neuroscience is mostly focused on biophysical models of neurons, neurodynamics and analysis of various brain signals. Good models of spiking neurons have been developed [31], and although mathematical characterization of their power has been described [32] their practical applications are still limited. The goal is to learn about the detailed mechanisms of biological systems, rather than try to abstract essential principles to make useful algorithms. Neuroinformatics is a new branch of science that is focused mostly on support of brain research, databases and tools for analysis of brain imaging and electrophysiological recordings. Neural ideas are also used in affective computing, trying to understand and express emotional content of speech and behavioral acts. Human-Computer Interaction (HCI) is a field that is mostly focused on modeling the cognitive limitations of human information processing. Neural models of higher cognitive functions are still very limited to small-scale experiments [33] and will certainly be quite complex. Neurocognitive informatics is aimed at abstracting the principles and creating large scale applications, therefore the overlap here is rather limited.

Fuzzy set theory has its inspirations in extensions of multivalued logics, but fuzzy logic and fuzzy rules may also be related to perception, object recognition, prototype-based categorization [34] and associative reasoning. These inspirations are more on psychological than neural level. Sets of fuzzy rules have a natural graphical representation [35], and are deeply connected to neural networks and uncertainty in measurements [36]. In particular if separable functions are used in neural network nodes their outputs may be treated as linear aggregation of fuzzy decisions based on product norms [16]. Fuzzy rules organized in a network form may also be tuned by adaptive techniques used in neural networks, therefore they are called neurofuzzy systems [37][38]. Neural networks may also use more sophisticated transfer functions, combining weighted activations of their inputs with distance-based activations [39][40]. In such a case each node does not model a single neuron but a larger group of neurons, perhaps a microcircuit, that performs more complicated function. Connectionist modeling in psychology [41] goes much further, assuming that network nodes represent whole concepts, and their connections represent explicit relations between these concepts. Each node in fact corresponds to many states of network subconfigurations that code concepts, although exact relations of connectionist models to neural processes have not yet been elucidated. Such networks may still incorporate the competitive learning "winner-takes-most" mechanisms. Semantic networks [42] simplify this further, simply linking the concepts with connections that designate relations of different types. Connectionist and semantic networks are examples of general graphical models [42], with other examples including Bayesian belief networks and abstract network models for parallel distributed processing.

Concept nodes, representing information derived from perception or from some measurements, are not too precise and thus may be represented in terms of fuzzy membership functions. Fuzzy models may be formally derived as generalization of multivalued logics, but the field has also rather obvious cognitive inspirations related to models of intelligent behavior at a higher, psychological level, rather than elementary neural level. Some efforts have been made to use fuzzy ideas in computing with words [44] and computing with perceptions [45], but these efforts are restricted to understanding the meaning of a small subset of linguistic phenomena and do not address NLP problems, or other cognitive systems problems. Granular computing becomes fashionable, defining "granules of information" at various level, and although on the mathematical side most algorithms used in this field can be related to clusterization one can imagine distant connections of information granules with activations of different brain structures. Fuzzy and neural systems are at the core of computational intelligence.

Computational intelligence community is also interested in statistical methods that grew up from research on perceptrons. Better understanding of mathematical foundations of neural techniques resulted in popular statistical pattern recognition models, such as the Support Vector Machines (SVMs) [46] for supervised learning, or Independent Component Analysis [47] and similar techniques for unsupervised learning. Probabilistic, Bayesian models, maximum likelihood and other approaches are being added to the bag of Computational Intelligence tools, but very little is being done with these methods in respect to knowledge representation, language or reasoning.

To cover all phenomena related to intelligence in a computational framework representation and reasoning based on complex knowledge structures is needed. Artificial intelligence (AI) has focused on symbolic knowledge representation, addressing problems that require reasoning, planning, understanding of language. These high-level cognition problems require quite different approach than problems that Computational Intelligence is addressing. Reasoning using linguistic concepts requires knowledge representation of diverse types of information that may be combined in an infinite number of ways. Logical, rule-based or frame-based representation schemes are quite useful but in many respect rather limited, inadequate to model associative processes that are always active in the brain. A conceptual blending theory of cognition [48] is based on assumption that various features and objects may be subconsciously "blended" facilitating creative thinking. This requires associative combinations of features of different representations. Although it is not clear how this process could proceed in the brain the theory has been partially implemented at the symbolic AI level [49]. It is certainly an important source of inspiration for neural models of creative processes and higher cognitive functions. Making inferences from partial observations and integrating them requires systematic search techniques and may draw inspirations from decision-making processes in which prefrontal cortex of the brain is involved.

One of the big challenges facing CI and AI communities is the integration of the algorithms that serve for modeling of both low and high-level cognitive processes. Autonomous robots need to reason using information based on perceptions, coming from sensors, categorized into information granules that are easier to handle. Although initial steps towards such integration have been made long time ago [50] many textbooks on artificial intelligence discuss only symbolic machine learning methods [3][4][51] that generate logical rules. Learning the rules directly from signals and measurements is certainly necessary in cognitive robotics and there are good algorithms to achieve this [30][52]. Recent integration of two major cognitive architectures, the neural Leabra [33] and the symbolic ACT-R, into one SAL architecture [53], showed that symbolic AI approaches may be understood as rough approximation to the brain functions. Such hybrid models are able to account for quite complex behavior and higher mental functions better than purely neural models.

Arguably the biggest challenges lie in the language domain. Language is the natural form of communication, both between people and to some degree within the human brain that strives for narrative understanding of its own history. Adding verbal comments to complex internals states helps to recall them, understand and reason about them. The ideal information management system should respond to queries in natural language. A natural "talking-head" interface is relatively easy to add [54] to chatterbots, but linguistic abilities of such programs are very limited. We shall focus here on neurocognitive inspirations in the language domain, although the goals of neurocognitive informatics are much wider. Providing simple models of higher cognitive functions should help to understand these functions better, and simplify them to create practical algorithms for large-scale applications. These functions include knowledge representation, different forms of memory (recognition, semantic, episodic, procedural, long and short-term, and working memory), connections of language to perception, thinking, problem solving, planning, attention, behavioral control, imagery, creativity and consciousness.

Although many details of processes in the brain are not known what we have learned so far is already a great source of inspiration. Intuition, insight, imagery, creativity and other brain functions are being used as inspirations for computational algorithms [55]. The connectome project [56] aims at filling the gaps in our knowledge about the information flow in the brain. The involvement of numerous cortical and subcortical brain areas in information processing is already partially known. This should allow for creation of brain-based representations of linguistic concepts. Other goals of neurocognitive informatics include models of spreading neural activation and priming effects in psychology, especially their influence on associative thinking; the transition from conscious to unconscious control in learning new skills; the process of learning a profession, from novice to an expert; understanding and modeling of various forms of imagery and its connections to creativity and talent; development of basic concepts in infant brains, understanding of affective states, and the effects of various dysfunctions at the low-level (neural ionic channels, degeneration) on mental functions.

For the first time in history the mind-body problem has solid scientific foundations based on correlations between brain states, as measured by its metabolic or electrical activity, and mental states (intentions, perceptions). The brain-mind transformation essentially requires transformation of neural activity $x_i(t)$, for example measured by $i=1..n$ EEG electrodes, into mental activity $m_j(t)$, where each dimension in the "mental space" is related to some quality of our inner experience, such as intentional feelings, tactile, auditory, visual and other sensory features.

3. **Large-scale NLP projects.**

This section presents inspirations for natural language processing based on current understanding of the neuroscience of language. The goal is to connect neural processes responsible for language processing with more traditional ways of analysis, such as semantic networks and vector models, in hope that this will lead to useful enhancement of existing models. J. Feldman in his book on a "Neural Theory of Language" (NTL) [57] stressed two points: "Thought is structured neural activity", and "Language is inextricable from thought and experience". Many current attempts to create large-scale semantic networks, ontologies, and lexical resources ignore these facts, trying to learn everything from statistical correlations that may be found in the text. Lexical resources, such as WordNet[1], has been described as "a lexical reference system whose design is inspired by current psycholinguistic theories of human lexical memory. English nouns, verbs, adjectives and adverbs are organized into synonym sets, each representing one underlying lexical concept. Different relations link the synonym sets" [58]. This is certainly a step in good direction, but the definitions and properties of concepts provided by Wordnet are very brief and do not allow to understand the meaning of such concepts. For example, "a ball" in one of the senses is defined as "a spherical object used as a plaything", and "plaything" is defined as "an artifact designed to be played with". A ball has many properties that we know from experience: size, color, softness or hardness, surface texture, some balls may be held in hand, or kicked, or bounced on the head, or hit with a stick. In fact "a ball" is a high level concept that subsumes many experiences. A detailed representation of many specific types of balls is stored in our brain and activated by the context in which this word appears. We know that the golf ball is smaller than the cricket ball, that a small steel ball is called a shot, and that in some languages (for example, Polish) a bullet is called a ball, even if it is not quite round. Even for such trivial example any human could build a whole ontology of different types of balls. The information in Wordnet, which is the largest hand-crafted project of its kind, with more than 200,000 words-sense pairs, is minimal, not sufficient to understand any concept, ask intelligent questions to get precise information about the concept, and use it meaningfully in a discussion.

In recent year many large-scale projects have been formulated, hoping to collect sufficient linguistic information for NLP systems. In the internet-based collaborative Open Mind Common Sense Project over 15,000 authors typed since the year 2000 over 700,000 sentences with all kind of obvious "common sense" facts. The ConceptNet knowledgebase derived from this project [59] has 200,000 assertions and the full base contains 1.6 million assertions. These assertions include spatial, physical, social, temporal, and psychological aspects of everyday life, capturing many commonsense concepts and relations in a large semantic network with over 150.000 nodes. The Commonsense Computing project at MediaLab (MIT) plans to use this knowledge and acquire more through collaborative projects. The Open Mind Commons[2] is an application designed to teach the system new information and refine existing knowledge. The project is growing and has also a databases in Portugese. The concept of "a ball" is characterized here a bit better than in Wordnet: a ball is round, circular, a sphere, is a toy, can bounce, roll, break window, is being thrown, is used in football, basketball, soccer, bowling, golf, snooker, sport, ... There are many statements like "A ball can sit on top of a VCR", or "You are likely to find a ball in the mouth of a puppy", which are not informative, and many contradictory statements: "the globe is a ball", "the globe is not a ball". Altogether in March 2009 there were 171 statements that mentioned the ball. This information is much richer than what is found in the Wordnet, but still contains only a small fraction of what ordinary people know about "the ball" concept.

Several interesting projects based on ConceptNet have been formulated. Mindpixel[3] project tried to create a knowledgebase by validating true/false statements and probabilistic propositions. Multi-Lingual version of ConceptNet semantic network has been used in LifeNet, based on concept relationships. It should learn from people who play simple online games, and also learn from sensors to perform activity recognition. The system should function as the first-person spatio-temporal reasoning engine, predicting and commenting events in a typical person's life. Although it looked very atractive the project has not progressed in recent years. It has also used result of the Honda Open Mind Indoor Common Sense project, based on the same principle, based on the same principle as ConceptNet, but focus on information that indoor mobile robots should know to have commonsense and recognize objects and indoor scenes. Examples given at the project webpage[4] are: "coffee is made in a coffee maker which is in a kitchen; to find out if it is raining one needs to look out of an open window; master bedroom usually has an attached bath".

---

[1] http://wordnet.princeton.edu
[2] http://commons.media.mit.edu/en/
[3] http://en.wikipedia.org/wiki/Mindpixel
[4] http://openmind.hri-us.com/login.jsp

Other important large-scale project that goes beyond WordNet ideas of a lexical database in which semantic relations, such as synonymy, antonymy, hyponymy, hypernymy and meronymy help to define meaning, is HowNet [60]. This project[5] has been in development for over 20 years. Hownet stresses not only ontological, but also temporal aspects of concepts, and has its own structure of inter-concept and inter-attribute relations. The idea of a sememes, smallest basic units of meaning, is rather natural in Chinese, as the ideographical characters used for writing refer to such units. Analysis of 6000 Chinese characters led to about 1600 sememes that have been used to define about 100.000 concepts for English, and roughly the same number of Chinese language. While WordNet provides barely enough information to distinguish between concepts HowNet is more constructive, adding not only superordinate terms (hypernyms), but many more specific dependency relations, for example attribute-host, material-product, event roles, dynamic role, concept co-occurrence, etc. The Knowledge Dictionary stores the main and the secondary features of concepts, synonyms, antonyms and converse relations, event-relatedness and role-shifting information, organized in supplementary databases. Still this database is very limited, with a single attribute for a ball, shape=round.

The FrameNet project[6], developed at Berkeley, creates a rich on-line lexical resource for English and Spanish, based on the frame semantics [61]. The third release (mid-2008) of this lexical database contained more than 10,000 lexical units (word-meaning pairs), over 60% fully annotated in more than 825 semantic frames and demonstrated in over 135,000 annotated sentences. This exercise is aimed at illustrating all valid ways of using each word in each of its senses. For example, the verb bake is linked to 3 frames: Apply_heat, Cooking_creation, and Absorb_heat, depending on the word usage. This approach has its roots in Case Grammars [62], replacing a small number of fixed roles with large number of semantic relations empirically derived from annotation of texts. FrameNet has focused more on verbs than nouns and therefore is not suitable as an ontology of things, but FrameNet has been linked to Suggested Upper-Merged Ontology (SUMO)[7]. FrameNet has developed a visualization tool for viewing the relations between frames. Connections between frames and their elements may be visualized. Automatic labeling of text with semantic frame information could greatly facilitate machine translation and semantic search, but so far are done mostly in manual way. This approach may be quite successful in specialized domains. A trilingual (English, German, French) lexical resource called The Kicktionary [63], describes the language of soccer. It combines frame semantics and WordNet style semantic relations, with around 2,000 lexical units organized in 104 frames and 16 scenarios, illustrated by multilingual corpus of soccer match reports.

The MindNet project at Microsoft Research[8] is aimed at building in a fully automatic way semantic networks from dictionaries, encyclopedias, and free text. Input sentences are parsed, semantic dependency graphs built, and many individual graphs aggregated into a single large graph. Probabilistic weights are assigned to subgraphs depending on their frequency in the corpus. This is an interesting way of acquiring world knowledge on the scale needed to support common-sense reasoning. Relations between concepts may be traced using an on-line browser that shows intermediate concepts, frequent collocations or similarity of paths in the subgraphs created by the system. MindNet has been used in machine translation projects at Microsoft but the last presentation has been in 2005.

These project represent large-scale efforts to collect linguistic data from free texts and structured sources of information, using automatic or labor-intensive algorithms. After a period of enthusiasm many projects slow down and despite brilliant ideas seem to come to a standstill. None of these projects has tried to understand how human brains implement language, how can we know so much by learning so little? The short answer is: language describes inner activation of brain areas, that is experience.

---

[5] http://www.keenage.com
[6] http://framenet.icsi.berkeley.edu/
[7] http://www.ontologyportal.org
[8] http://research.microsoft.com/en-us/projects/mindnet/default.aspx

## 4. Language and memory

We intend to approach NLP using inspirations from general understanding of the human information processing, investigate the feasibility of text understanding based on neurocognitive inspirations, namely representation of words in the brain and the use of various brain memory systems. Neurocognitive approach to NLP follows inspirations from brain science focusing on approximated models of memory and other neural processes. The long-term goal is to reach human-level competence in natural language processing.

Schank was right in his early "conceptual dependency" theory of understanding language [64] claiming that language cannot be studied without taking memory into consideration. Analysis of texts, independent of the purpose, requires three main steps, that include at least:

- mapping from strings of letters to unique terms, recognition of tokens;
- grouping terms into phrases, mapping them to concepts and resolving ambiguities;
- semantic representation of the whole text, capturing relations among entities that are involved, facilitating inferences, and thus allowing for understanding and answering questions about its content.

These three steps roughly correspond to functions of four kinds of human memory [65]: recognition, semantic, episodic and working memory. Semantic networks [42] have been explicitly inspired by the theory of semantic memory, but NLP research largely ignored these inspirations, focusing on formal approaches [66].

**Recognition memory** helps to ignore most spelling errors. Context and anticipation plays a major role in correct recognition. As long as the first and the last letter of the word is not changed even severely distorted texts containing *wrods wtih many paris of letres trasnpoesd* is read without much trouble, a phenomenon that is of interest to spammers and cognitive scientists. Unstructured free texts, and especially conversational texts need a lot of data cleaning. This is traditionally done by spellcheckers, but without understanding the context may lead to wrong recommendations. For example, Lexical Systems Group of the US National Library of Medicine has developed a spelling suggestion tool called Gspell[9] that is based on the SPECIALIST lexicon containing general English words and many biomedical terms. Medical terminology is difficult to spell for the average English speaker ("14% of all queries submitted for health information retrieval contain a misspelled term", [67]). Gspell does not use context to understand the topic and makes many spelling suggestions, although humans "see" only a single term, frequently paying no attention to the misspellings. Reading text leads to priming effects in the brain, creating expectations and anticipations for a few selected words, and inhibition of many others that "do not come to our mind". Recognition memory cannot be separated from other memory systems, doing much more than just searching for similar terms in the lexicon.

According to Hecht-Nielsen's confabulation theory of the brain functions [68] generating expectations is almost all that neocortex does. Statistical language processing models applied to a very large text corpus used for training allowed him for prediction of the next word in a sequence with high reliability, partially capturing this anticipation, although statistical algorithms do not approximate well real brain processes behind this phenomenon. Anticipation may help to disambiguate word senses, facilitating the mapping of terms into concepts. Although his model has not been directly used for spell-checking that includes context it will lead to a quite similar approach to that taken by Google in "Google suggest", that is based on statistics of longer n-grams. Local context is also included in another tool also called Gspell (for MacOS only), a spelling checker that searches Google to find spelling recommendations, including personal names, site names and other words that are not found in standard dictionary. A better approximation to functions of biological recognition memory should create possible candidate words that could result from various misspellings, and remove (inhibit) those that do not fit to the current context, defining the topic of discourse, or an episode. This will be elaborated further below.

---

[9] http://lexsrv3.nlm.nih.gov/SPECIALIST/Projects/gSpell/V0.0.40/GSpell.html

**Semantic memory** encodes in the activity of brain's subnetworks factual information about objects and concepts, together with their properties and relations, general knowledge of the world largely independent of context and personal relevance. It is thus an interesting inspiration for knowledge representation. Formal models of semantic networks inspired by psychological ideas about semantic memory [65],[70],[72] have been developed in the last three decades in artificial intelligence [42],[69]. These models were created at the time when not much has been known about neurobiology of semantic memory and thus do not try to approximate functions of biological semantic memory (SM). Psychologist simply postulated nodes that correspond to concepts and different type of links that connect them, without thinking about the actual implementation of these processes in the brain. Such structure of semantic network does not take into account the fact that each node in such network is in fact a neural circuit, with similarities and associations between concepts resulting from sharing some common elements or mutual activations that are responsible for semantic priming.

The *semantic priming* (SP) phenomenon has been known in cognitive psychology since more than 30 years, and has been extensively studied in experimental psychology (see the review by McNamara [71]). Each word excites brain subnetworks that encode different meanings of that word [5]. In such coding identical phonological representations of words may be shared among several concepts without leading to any problems. Words that have been processed earlier (context) have already activated many brain subnetworks, increasing the probability of a particular meaning of the new concept, and inhibiting all other meanings. This competition, leading to inhibition of subnetworks coding alternative meanings of the word, makes it hard to think about alternatives when one of the meanings (interpretations) fits really well to the current context. These alternatives simply "do not come to our minds". Semantic memory develops slowly as a result of learning, forming a view of the world that reflects frequently repeated situations and regularities. Training neural networks associations between facts, description of properties of various objects, captures similarity of these objects in the activations of hidden neurons (corresponding to association cortex, connected to sensory and motor areas). So far this process has been modeled only on a toy example [72],[73] but may provide a bridge towards practical NLP algorithms based on vector representations of the activations of different parts of the brain. Recent results with fMRI neuroimaging of brain states [74] show that when the subject thinks, hears or sees an image invoking an object similar activity of the brain arises even in different people. Prototype probability distributions of activation of different brain areas may thus serve as a brain-based vector representation that approximates this process.

Semantic memory is frequently not sufficient for disambiguation of concept meanings because a wider context may be needed. This is provided by the **episodic memory** that includes times, places, associated emotions, and other contextual knowledge of autobiographical events, incorporating also current situation, relations between objects, participants, understanding the topic of the text fragments and discourses. In short, this type of memory encodes who did what to whom where and when. Episodic level of encoding and processing information facilitates selection of interpretations at the semantic and recognition level. All three levels are in fact strongly coupled, because all brain areas are strongly coupled, creating in a fraction of a second one dynamical state. A few computational models of general episodic memory exist [75]-[77], with many others focused on the role of hippocampus [78]. In the TraceLink model [75] hippocampus provides links that activate cortical columns, forming coherent quasi-stable attractors corresponding to episodic states. These links may serve as episodic memory representation, an alternative to distribution of activations over columns or larger regions of the brain.

**Working memory** also plays important role in forming episodic memory. Initially it has been postulated as a theoretical psychological construct by Baddeley and Hitch [80], serving as a gateway between sensory input long-term memory. Neuroimaging experiments and lesions studies showed involvement of prefrontal and parietal cortex, anterior cingulate, and parts of the basal ganglia. This is a rather complicated network of diverse structures, actively involved in learning skills, providing temporal storage for manipulation of information and storing relevant information permanently [81][82]. Part of the content of working memory is experienced in a conscious way and plays key role in the reinforcement learning of skills [83]. Biologically-inspired cognitive architectures, such as Leabra [33] provide models of prefrontal cortex capable of short term storage that is preserved despite on-going short-term memory activity. ACT-R and other symbolic architectures treat working memory simply as a collection of elements (rules, conditions) that have actually been recalled and are available for use [84].

Different knowledge representation schemes suitable for different applications may be formulated starting from such inspirations. Creation of large-scale semantic networks, and coupling with models of episodic memory for topic categorization and recognition memory for expansion of abbreviations, acronyms and misspellings is going to be of fundamental importance in all NLP-related areas. However, more detailed analysis of language-related brain activity will be even more inspiring.

## 5. Words in the brain and the role of non-dominant hemisphere

How are words and concepts represented by brains? Acoustic speech input is quickly changed into categorical, phonological representation. This representation helps to understand language in a voice-independent way, but also make understanding of languages that have quite different phonological structure from our native language rather difficult [85]. A small set of phonemes is linked together in ordered string by a resonant state representing word form, and extended to include other brain circuits defining semantic concept. Hearing a word phonological processing creates localized attractor state whose activity spreads in about 90 ms to extended areas defining its semantics [5]. To recognize a word in a conscious way activity of its subnetwork must win a competition for an access to the working memory [86]-[87]. Hearing a word activates strings of phonemes priming (decreasing the threshold for activity) all candidate words, as well as some non-word phoneme combinations. Polysemic words may use a single phonological representation that differs only by semantic extension. Unless attention is paid to the form of the word (decreasing activation thresholds for phonological representations), as when puns are deliberately made, there is little interference from other meaning of the word thanks to the inhibitory processes. This point of view agrees with the cell assembly model of language that has already quite strong experimental support and agrees with broader mechanisms responsible for memory. In the cell assembly (or neural clique) model words (or general memory patterns) are represented by strongly linked subnetworks of microcircuits that bind articulatory and acoustic representations of spoken words. The meaning of the word comes from extended network that binds related perceptions and actions, activating sensory, motor and premotor cortices. Various neuroimaging techniques confirm existence of such semantically extended networks.

In people who can read and write visual representation of words in the recently discovered Visual Word Form Area (VWFA) in the left occipitotemporal sulcus is strictly unimodal [86]. Adjacent Lateral Inferotemporal Multimodal Area (LIMA) reacts to both auditory and visual stimulation and has cross-modal phonemic and lexical links [89]. It is still controversial, but the auditory word form area may also exists [86][88]. It may be a homolog of the VWFA in the auditory stream, located in the left anterior superior temporal sulcus; this area shows reduced activity in developmental dyslexics. Such representation of words should help to focus symbolic thinking. Context priming selects extended subnetwork corresponding to a unique word meaning, while competition and inhibition in the winner-takes-all processes leaves only the most active candidate network. Semantic and phonological similarities between words should lead to similar patterns of brain activations for these words.

Language cannot be separated from memory and thinking processes. In the dominant, usually left hemisphere (LH), phonological representations may encode precisely word forms in local circuits capable of strong activations (as differences between words and non-words in evoked response potentials [5] seem to indicate), and thus quickly activate extended networks that make words meaningful. The role of contralateral hemisphere, usually the right one (RH), in language processing is usually reduced to analysis of prosody and non-verbal cues, but there is growing evidence that it is also involved in language interpretation [90]. Additional information on the role of right hemisphere comes from studies of the sudden insight (Aha!) experience that accompanies solutions of some problems.

In experiments with functional MRI and EEG problem solving situations that have been accompanied by insight have been contrasted with those when step-by-step gradual solution did not required insight [91],[92]. About 300 ms before the moment in which Aha! experience was felt a burst of gamma activity was observed in the Right Hemisphere anterior Superior Temporal Gyrus (RH-aSTG). The delay of 300 ms is consistent with studies showing delay of subjectively felt conscious experience in respect to the brain activity [93]. The gamma burst has been interpreted as "making connections across distantly related information during comprehension (…) that allow them to see connections that previously eluded them" [91]. Bowden *et al.* [92] performed a series of experiments that confirmed the EEG results using fMRI techniques. One can conjecture that this area is involved in higher-level abstractions that can facilitate indirect associations [95]. The initial impasse in problem solving is due to the inability of the processes in the left hemisphere, focused on the precise representation of the problem, to make progress in finding a solution. Solution means that from initial configuration $\mathbf{S}_0$ of neural activations representing the problem, a series of transitions states $\mathbf{S}_k$, each with phonological component (symbolic interpretation), may be generated, until desired configuration $\mathbf{F}=\mathbf{S}_n$ is reached. The requirement for symbolic interpretation means that intuitive solutions, when we feel that some conjecture is true but are not able to justify it, may be correct but are not sufficient.

The RH has only an imprecise view of the left hemisphere (LH) activity, generalizing over similar concepts and their relations. This activity represents abstract concepts, corresponding to categories higher in ontology, but also captures complex relations among concepts, relations that may not have symbolic name, but are useful in reasoning and understanding. Sometimes we feel that something is quite out of context, or that the sentence we are producing is hard to complete in grammatical way, and this may be reaction of language-related areas in the right hemisphere. For example, "left kidney" sounds correct, but "left nose" seems strange, although we do not have a concept for spatially extended things that "left" applies to. The feeling arising from understanding words and sentences may be connected to the left-right hemisphere activation interplay. Most of RH activations do not have phonological components; the activations result from diverse associations, temporal dependencies and statistical correlations that create certain expectations. Current view of recognition memory distinguishes two functionally distinct processes, recollection and familiarity (dual process theory [96]), giving rise to distinct phenomenal states. Familiarity-based memory leads to a feeling of knowing without reference to specific information. Conclusion of a recent paper [97] where familiarity with words presented 10 minutes before was tested are: "Across a variety of brain regions, the neural signature of recollection was found to be distinct from familiarity, demonstrating that recollection cannot be attributed to familiarity strength". The effects in the left hemisphere were stronger, "but in most cases, similar but weaker activity was observed in the right hemisphere". One can assume that interpretation of speech or text is greatly enhanced by "large receptive fields" in the RH, which can constrain possible interpretations, help in the disambiguation of concepts and provide ample stereotypes and prototypes that generate various expectations.

Impasse in problem solving may be overcome by decreasing activations in the verbal language areas (when the conscious efforts to solve the problem are given up) and trying to solve the problem at the non-verbal, intuitive level. Activations of larger subnetworks in the RH may be sufficient to connect the starting and the desired final configurations $\mathbf{S}_0$ and $\mathbf{F}$, creating a strong resonance state that will give rise to a gamma burst. Sometimes it leads to a brief feeling that the solution is imminent, although it has not yet been formulated in symbolic terms, a common feeling among scientist and mathematicians. The LH impasse is removed when relevant activations are projected back from the less-focused right hemisphere, allowing new dynamical associations to be formed among brain subnetworks primed by the problem description. High-activity gamma burst is projected to the left hemisphere prime LH subnetworks with sufficient strength to activate intermediate configurations, making the transition from $\mathbf{S}_0$ through $\mathbf{S}_k$ to $\mathbf{F}$ possi-ble, creating associative connections linking the problem statement with partial or final solution. An emotional component is needed to increase the plasticity of the brain and remember these associations.

The "Aha!" experience may thus result from the activation of the left hemisphere areas by the right hemisphere, with a gamma burst helping to bring relevant facts to the working memory, making them available for conscious processing. The final step in problem solving requires synchronization between several left hemisphere states, representing transitions from the start to the goal through intermediate states. This seems to be a universal mechanism that should operate not only in solving difficult problems, but also on much shorter time scale in understanding of complex sentences. The feeling "I understand" signifies the end of the processing and readiness of the brain to receive more information.

Sigmund Freud is frequently quoted: "When making a decision of minor importance, I have always found it advantageous to consider all the pros and cons. In vital matters, however, the decision should come from the unconscious, from somewhere within ourselves.". The Nijmegen Unconscious laboratory[10] at Radboud University focuses on issues unconscious thought, and their results agree with this statement, pointing to the important role of the right hemisphere in thinking and decision making.

### 6. Creativity in language comprehension and production

Understanding language is not only a process based on logical inferences but requires also imagination and creativity. For example, hearing new words or novel expressions that cannot be found in the dictionary we may usually guess their sense. Producing non-trivial sentences, surprising analogies or figures of speech may also be a creative act. Creativity research is done mostly by psychologist, with a few exceptions [98], where EEG studies have shown greater neural activity in the right parieto-temporal areas, higher alpha activity during impasse or inspiration period, and greater tendency for emotional excitement, all of which agrees with remarks on the role of right hemisphere presented in the previous section. New results obtained have been reported only very recently by the "positive neuroscience" group [99] that studies creativity using multimodal imaging techniques. A network of brain regions involved in creative thinking has been identified, comprised of the anterior cingulate gyrus, superior temporal gyrus, and the corpus callosum.

Relationships between creativity and associative memory processes have been noticed a long time ago [100]. The pairwise word association technique is perhaps the most direct way to analyze associations between subnetworks coding different concepts. These associations should differ depending on the type of priming (semantic or phonological), structure of the network coding concepts, the activity arousal due to the priming (the amount of energy pumped into the resonant system). In a series of experiments [101] phonological (distorted spelling) and semantic priming was applied, showing for a brief (200 ms) moment the priming cue (word) before the second word of the pair was displayed. Two groups of people, with high and low scores in creativity tests were participating in this experiment. Two type of associations were presented, simple (close) and difficult (remote), and two types of priming, positive (either phonological or semantic relation to the second word) and neutral (no relation). Creative people should have greater ability to associate words and should be more susceptible to priming. Less creative people may not be able to make remote associations at all, while creative people should show longer latency times before noticing such associations or claiming their absence. This is indeed observed, but other results have been quite puzzling [101].

Neutral priming, based on the nonsensical or unrelated words, increased the number of claims that words are related, in case of less creative people stronger than positive priming, and in case of more creative people in a slightly lower way. Phonological priming with nonsensical sounds partially activates many words, adding intermediate active configurations that facilitate associations. If associations between close concepts are weak neutral priming may activate intermediate neural oscillators (pumping energy to the system, increasing blood supply), and that should help to establish links between paired words, while positive priming activates only the subnetwork close to the second word, but not the intermediate configu-rations. For creative people close associations are easy to notice and thus adding neutral or positive primes has similar small effect. Situation is quite different for remote associations. Adding neutral priming is not sufficient to facilitate connections in less creative brains when distal connection are completely absent, therefore neutral priming may only make them more confused. Adding some neural noise may increase the chance to form resonance state if weak connections exist in more creative brains – in the dynamical systems language this is called the stochastic resonance phenomenon [102]. On the other hand adding positive priming based on spelling activates only phonological representations close to that of the second word, therefore there is no influence. Priming on positive (related) meaning leads to much wider activation, facilitating associations. These results support the idea that creativity relies on the associative memory, and in particular on the ability to link together distant concepts.

---

[10] http://www.unconsciouslab.com/index.php

Problems that require creative solutions are difficult to solve because neural circuits representing object features and variables that characterize the problem have only weak connections, and the probability of forming appropriate sequence of cortical activities is very small. The preparatory period – reading and learning about the problem – introduces all relevant information, activating corresponding neural circuits in the language areas of the dominant temporal lobe, and recruiting other circuits in the visual, auditory, somatosensory and motor areas used in extended representations. These brain subnetworks are now "primed", and being highly active reinforce mutually their activity, forming many transient configurations and inhibiting at the same time other activations. Difficult problems require long incubation periods that may be followed by an impasse and despair period, when inhibitory activity lowers activity of primed circuits, allowing for recruitment of new circuits that may help to solve the problem. In the incubation period distributed sustained activity among primed circuits leads to various transient associations, most of them short-lived and immediately forgotten. Almost all of these activations do not have much sense and are transient configurations, fleeting thoughts that escape the mind without being noticed. This is usually called imagination. Interesting associations are noticed by the central executive and amplified by emotional filters that provides neurotransmitters increasing the plasticity of the circuits involved and forming new associations, pathways in the conceptual space.

Results of experimental and theoretical research lead to the following conclusion: creativity involves neural processes that are realized in the space of neural activities reflecting relations in some domain (in case of words knowledge about morphological structures), with two essential components: 1) distributed chaotic (fluctuating) neural activity constrained by the strength of associations between subnetworks coding different words or concepts, responsible for imagination, and 2) filtering of interesting results, amplifying certain associations, discovering partial solutions that may be useful in view of the set goals. Filtering is based on priming expectations, forming associations, arousing emotions, and in case of linguistic competence on phonological and semantic density around words that are spontaneously created (density of similar active configurations representing words).

It is thus quite likely that language comprehension and creative processes both require subsymbolic models of neural processes realized in the space of neural activities, reflecting relations in some experiential domain, and therefore cannot be modeled using semantic networks with nodes representing whole concepts, nor other techniques, that have been used in the large-scale projects described above.

## 7. Neurocognitive informatics approach to language

Neurolinguistic observations described above may be used as inspirations for neurocognitive models. The same neural processes should be involved in sentence understanding, problem solving and creative thinking. In the past there have been only a few attempts to use inspiration from brain research in NLP. Lamb defined the goal of neurocognitive linguistics as: "an attempt to understand the linguistic system of the human brain, the system that makes it possible for us to speak and write, to understand speech and writing, to think using language …" [103]. Although his approach has been quite interesting and fruitful in understanding the neuropsychological language-related problems it had no ambition to lead to creation of algorithms for text interpretation, and it is still quite exotic in the natural language processing (NLP) community. The basic premise is rather simple: each word in analyzed text is a part of an associative net-work where activation spreads and the states of the networks facilitate semantic interpretation of the text, yet unraveling these "pathways of the brain" is in practice quite difficult.

Connectionist approach to natural language has been introduced in the influential PDP books [41]. Miikkulainen created DISCERN [104][105], a system for subsymbolic natural language processing, with scripts, lexicon, and memory based on Kohonen's self-organizing networks [24], but despite very promising start abandoned this project. Application of constrained spreading activation techniques in information retrieval [104], semantic search techniques [107] and word sense disambiguation [108] have recently been made. Experience so far shows that direct attempts for large-scale neural network applications to language problems may be rather difficult and therefore a more promising direction would be to find approximations of brain activity that could be connected to the vector models, extensively used in statistical approaches to language [66]. Below a sketch of neurolinguistic inspirations that can be used to find useful approximations to spreading of brain activity during text comprehension is given.

Words activate several subnetworks, two of which will be distinguished here: one encoding the form of the word (phonological, but it may also be related to the written form in iconographic or alphabetic languages), and the second, extended over the whole brain, providing the meaning for that word. Thus approximation of the spreading activation in the brain during language processing should require at least two networks activating each other. Given the word $w = (w_f, w_s)$ with phonological (or visual written) component $w_f$, an extended semantic representation $w_s$, and the context *Cont* (previous information that has already primed the network), the meaning of the word results from spreading activation in the left semantic network *LH* coupled with the right semantic network *RH*, establishing a global quasi-stationary state $\Psi(w, Cont)$. This state rapidly changes with each new word received in sequence, with quasi-stationary states formed after each sentence is understood. It is quite difficult to decompose the $\Psi(w, Cont)$ state into components, because the semantic representation $w_s$ is strongly modified by the context. The state $\Psi(w, Cont)$ may be regarded as a quasi-stationary wave, with its core component centered on the phonological/visual brain activations $w_f$ and with quite variable extended representation $w_s$. As a result the same word in a different sentence creates quite different states of activation, and the lexicographical meaning of the word may be only an approximation of an almost continuous variation of this meaning. To relate states $\Psi(w, Cont)$ to lexicographical meanings, one can clusterize all such states for a given word in different contexts and define prototypes $\Psi(w_k, Cont)$ for different meanings $w_k$. Each word in a specific meaning is thus implemented as a roughly defined extended configuration of activations of a very dense neural network, with associations between the words resulting from transition probabilities from one word to the other, that follow from the connectivity and priming of the network. This process is rather difficult to approximate using typical knowledge representation techniques, such as connectionist models, semantic networks, frames or probabilistic networks.

The high-dimensional vector model of language popular in statistical approach to natural language processing [66] is a very crude approximation that does not reflect essential properties of the perception-action-naming activity of the brain [5][86]. The process of understanding words (spoken or read) starts from activation of the phonological or grapheme representations that stimulate networks containing prior knowledge used for disambiguation of meanings. This continuous process may be approximated through a series of snapshots of microcircuit activations $\phi_i(w, Cont)$ that may be treated as basis functions for the expansion of the state $\Psi(w, Cont) = \Sigma_i \alpha_i \phi_i(w, Cont)$, where the summation extends over all microcircuits that show significant activity resulting from presentation of the word $w$. The high-dimensional vector model used in NLP measures only the co-occurrence of words $V_{ij} = \langle V(w_i), V(w_j) \rangle$ in some window, averaged over all contexts. A better approximation of the brain processes involved in understanding words should be based on the time-dependent overlap between states $\langle \Psi(w_1, Cont) | \Psi(w_2, Cont) \rangle = \Sigma_{ij} \alpha_i \alpha_j \langle \phi_i(w_1, Cont) | \phi_j(w_2, Cont) \rangle$. Systematic study of transformations between the two bases: activation of microcircuits $\phi_i$ and activation of complex patterns $V(w_i)$, has not yet been done for linguistic representations in the brain. Analysis of memory formation in mice hippocampus [87] in terms of combinatorial binary codes signifying activity of neuronal cliques goes in this direction. The use of wave-like representation in terms of basis functions to describe neural states makes this formalism similar to that used in quantum mechanics, although no real quantum effects are implied here.

Spreading activation in semantic networks should provide enhanced representations that involve concepts not found directly in the text. Approximations of this process are of great practical and theoretical interest. The model should reflect activations of various concepts in the brain of an expert reading such texts. A few crude approximations to this process may be defined. First, semantic networks that capture many types of relations among different meanings of words and expressions may provide space on which words are projected and activation spread. Each node $w$ in the semantic network represents the whole state $\Psi(w,Cont)$ with various contexts clustered, leading to a collection of links to other concepts found in the same cluster that capture the particular meaning of the concept. Usually only the main differences among the meanings of the words with the same phonological representation are represented in semantic networks (meanings listed in thesauruses), but the fine granularity of the meanings resulting from different contexts may be captured in the clusterization process and can be related to the weights of connections in semantic networks. The spreading activation process should involve excitation and inhibition, and "the winner takes most" processes. Models of semantic networks used in NLP are only vaguely inspired by the associative processes in the brain and do not capture such details [42],[66],[69].

Crude approximation to the spreading activation processes leads to an enhancement of the initial text being analyzed by adding new concepts linked by episodic, semantic or hierarchical ontological relations. The winner-takes-all processes lead to inhibition of all but one concept that has the same phonological word form. Locally this may be represented as a subnetwork (graph) of consistent concepts centered around a prototype for a given word meaning $\Psi(w_k,Cont)$, linking it to words in the context. Such approach has been applied recently to disambiguate concepts in medical domain [95]. The enhanced representations are very useful in document clusterization and categorization, as is illustrated using short medical texts described in the next section. Vector models may be related to semantic networks by looking at snapshots of the activation of nodes after several steps of spreading the initial activations through the network. In view of the remarks about the role of the right hemisphere, larger "receptive fields" in the linguistic domain should be defined and used to enhance text representations. This is much more difficult because many of these processes have no phonological component and thus have representations that are less constrained and have no directly identifiable meaning. Internal representations formed by neural networks are also not meaningful to us, as only the final result of information processing or decision making can be interpreted in symbolic terms (this idea can be used to model creativity at the phoneme-word level [109][110]).

Results from fMRI studies of the brain activity associated with word meaning [74] cannot be used directly to create representations of concepts for practical applications. However, knowing a few experimental results and using a large corpus of words it has been possible to predict neural activity for many more words. Another approach is to list all important brain area, and try to evaluate how strongly a given word may activate them. This is now being done with emotional words, with dictionaries that label each word with several dimensions of valence. This may be very much extended, enumerating different brain areas and their functions and evaluating, in an automatic way, how strongly is a given concept associated with specific type of information processing. This should allow to create high-dimensional vector-based representations.

Surprising efficiency of latent semantic analysis (LSA) [112][113], a simple procedure that decomposes term-document matrix into eigenvectors, has been summarized by Landauer: "The lesson I then take from LSA's successes is that empirical association data, when sufficient to accurately induce how all of its elements are related to each other, makes learning from experience powerful enough to accomplish much, if not all, of what it does." Context relations may partially be captured by eigenvectors, and given sufficiently large corpus various wordforms, past and present verbs, singular and plural nouns in all forms (inflec-tions, derivations, and compounding) will be represented as closely related. However, LSA cannot be used for analysis of novel words or novel meanings generated by morphological composition, which are quite common in some languages. Antonyms are usually represented as highly similar. Also LSA does not provide universal representations of words, that we are hoping to create with brain-based approach.

Other spectral techniques that are related to LSA include probabilistic latent semantic analysis, non-negative matrix factorization, and latent semantic mapping.

## 8. Preliminary results

Ideas outlined above have been used in several projects. The key problem is to find the pathways of spreading neural activation (associations) in human brain during linguistic task. The texts or information in a discourse does not provide sufficient basis for understanding. Semantic memory that captures common sense and the knowledge of an expert should be the basis of such algorithms. Relations stored in semantic memory should be dynamically weighted, depending on the priming effects by the context, thus changing the activation flow in different situations. On the formal level this may be done within the vector model or probabilistic models by changing the weights in the similarity functions, depending on the overall evaluation of the contexts, including emotional cues, goals etc.

One simple algorithm that has been tested with very good results follows the relations that are stored in ontologies, and uses feature selection techniques to learn which relations are useful for discrimination. This is possible if many types of relations for a given concept are given. In medical domain the UMLS collection of ontologies [114], including the Medical Subject Headings (MeSH) [115], provides rich information, with many semantic types and relations between concepts. The algorithm requires a set of labeled documents – we have used summary discharges from hospital, that may be labeled by the main and the secondary disease names, but Wikipedia articles that are labeled by the subject headings may be used as training texts for other fields. The goal is to create vectors that will indicate which concepts should be added to the texts to make the meaning explicit. The algorithm for the medical texts proceeds as follows:

1. Perform the text pre-processing steps: stemming, stop-list, spell-checking, either correcting or removing strings that are not recognized.
2. Use MetaMap software [116] for concept discovery, with restrictive settings to avoid highly ambiguous results when mapping text to UMLS ontology, expand acronyms.
3. Use UMLS relations to create first-order cosets (sets of new concepts associated with concepts found in the text that help to constraint its meaning); add only those types of relations that lead to improvement of classification results.
4. Reduce dimensionality of the first-order coset space using feature ranking methods [117], but do not remove original (zero-order) features;.
5. Repeat steps 3 and 4 iteratively to create second- and higher-order enhanced spaces.
6. Create $\mathbf{X}(t_i)$ vectors representing concepts.

As a result initial texts are enriched by adding those

Vectors $\mathbf{X}(t_j)$ representing terms $t_j$ have zero elements except for $\mathbf{X}(t_k)=1$ for $k=j$ and for those terms that are in the cosets for a given term. They are highly dimensional and may be normalized to the unit length $\|\mathbf{X}(t_k)\|=1$ without loss of information; any metric may then be used to compare them. They contain important terms to help distinguish documents of different categories. The non-zero coefficients of these vectors show connections between related terms. Iterative character of the algorithm leads to non-linear effects, feedback loops are strengthening some connections. In medical document categorization a single specific occurrence of a concept may be an important indicator of the document category. The Latent Semantic Analysis will miss it, finding linear combinations of terms that do not have clear semantics.

This general algorithm helps to discover associations that are important from categorization point of view, and thus understanding of the text topics, or forming an episode. It helps to build semantic network with useful associations and has been used in many medical applications [118]-[124]. As a results short documents, such as hospital discharge summaries, in the enhanced space show strong clusterization, and concepts that appear in them are disambiguated without problems. Figure below shows how the similarity of documents, visualized using multidimensional scaling (using Ghostminer software[11]), looks like after the first 4 iterations of this procedure. Interesting clinical types may be identified, as some clusters show a mixture of several labels [118] (see Fig, bleow).

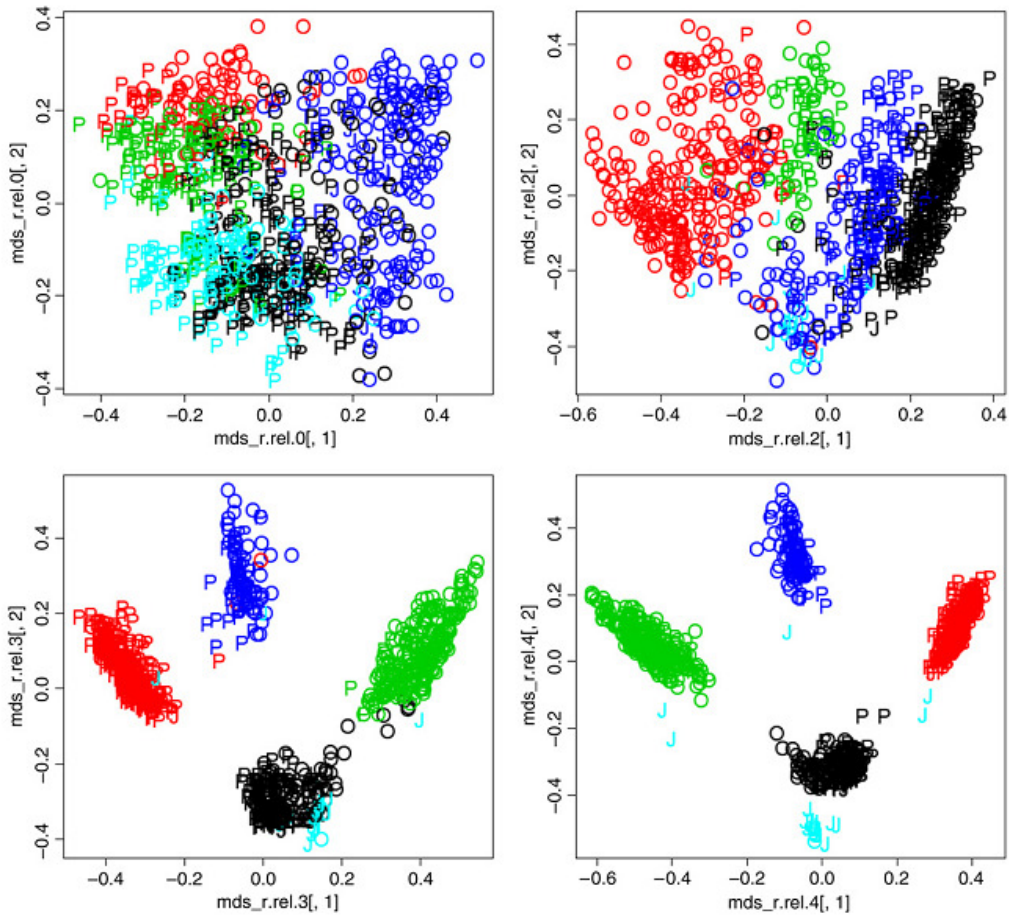

This process may also be used to follow the development of expertise in a given field. If a medical system is trained on general texts instead of real cases clusterization will not occur as these texts contain all possible combinations of concepts, and thus do not reflect probability of distributions of concepts that coherently describe observations that are encountered in a real world. Moreover, transition from a system that has learned from textbooks to a system that is an expert is slower than direct learning on properly selected cases with their clinical description. One may conclude that the current learning practices in medicine and many other fields are harmful, slowing the process of becoming an expert, therefore it takes many years after the study to gain and improve "intuitive" knowledge. Investigation of such processes using neurocognitive models of language is of great importance to education.

A knowledge-based algorithm to create synsets and replace word concepts by sense concepts is based on the assumption that given a reference corpus with sufficient text material on a given topic the same concept senses will be expressed in almost all possible ways. In such case reading this reference corpus will activate appropriate senses linked to concepts more often than the wrong senses. The algorithm has

---

[11] http://www. fqs.pl/ghostminer/

following steps:
1. Given a text $T$ create a corpus of reference texts $R$ on a similar topic; if there is no such corpus (textbooks, encyclopedia articles) find sufficient number of similar texts to $T$ and use it as the reference corpus.
2. Initial synset list $L$ has count $N(S)=0$ for all synsets $S$ (is empty); it will be used to prepare synsets for $T$.
3. Read reference texts $R$, take next word $u$ and check for collocations, creating concepts $w$.
4. Using Wordnet find all synsets $S_i(w)$, that contain $w$.
5. Add these synsets to the list $L$, increasing the count $N(S)$ by 1.
6. Keep reading until the end of the reference texts, creating a list of concepts $w$ linked to synsets $S_i(w)$.
7. Sort concepts and if more than one synset is linked to some concepts select the one with higher count number.

All concepts from the reference texts will have synsets and some new concepts may also appear, arising from reference concepts found in synsets. Reading the text $T$ to be annotated it is sufficient now to change all concepts to synsets. This algorithm may fail only in rare cases, when the same concept is used in the text in several meanings. Analysis of context is then necessary to distinguish these meanings. However, this happens only rarely and in many applications is not a problem at all. The quality of this approach is very high if appropriate reference texts are selected.

Another line of research that is of great importance is in semantic memory acquisition. In medical domain this could be done using UMLS and huge ontologies, but for general texts comparable resources do not exist. Although large-scale projects such as the ConceptNet may collect some useful facts some common-sense knowledge is very hard to collect in this way. For example, structural properties of physical objects are quite obvious to anyone capable of seeing, and therefore are never described. One way to acquire useful knowledge is to learn how to ask questions to discriminate between objects. The 20-question game may be used as a paradigm for gaining more information in case of ambiguities that cannot be resolved automatically, as well as in case of images that should be classified, or objects that humans can describe but do not know how to characterize. It is a form of active learning. The question/answers competitions organized so far were focused on searching for answers, while equally important part, formulating intelligent questions, has been neglected.

An implementation of the 20-question game has been used for many years[12] and is sold in electronic toys. This implementation is based on about 500 fixed questions that have been weighted depending on their usefulness in answering questions using results of past games. It is quite successful, pointing to the fact that even in a rather small feature space it is possible to distinguish between concepts. This version is quite limited as in practice we are given some information in the text and cannot start asking from the beginning in hierarchical way. A proper description of concept properties may, however, be sufficient to formulate questions that bring maximum information in a given situation. For example, if there are 10 candidate words for translation a series of questions to disambiguate them may be asked, and either answered by checking the context, or answered by humans.

A challenge is thus to create an algorithm that asks intelligent questions using semantic memory for concept description. Automatic ways to collect information for semantic memory, active search for plausible information that may be related to a given concept, correct it using dialogues, and finally use it to formulate questions, play the game and identify objects has been described in a series of papers [125]-[131].

At the sub-symbolic level one may also consider the process of understanding and creation of novel words. The basic assumption is that creativity in this domain requires imagination, which is based on spontaneous combination of morphological and phonological constituents of the priming words con-strained by probability distributions for different combinations of these constituents in a given language (learned from a corpus data), and filtering of interesting and novel combinations, depending on their phonological as well as associative strength – words that contain morphemes that link the m to many other words are considered to be interesting. Results of an algorithm based on these principles for creation of novel words [109] show that about 2/3 of the words have already been invented by people and are in use as the names of companies, products or websites, but about 1/3 is equally interesting but novel. These results give hope that also creative process in the brain can be approximated using neurocognitive ideas.

---

[12] http://www.20q.net

### 9. How to develop neurocognitive linguistics

All that was said above shows that looking at the brain for inspirations may be quite fruitful for NLP, if suitable approximations are defined. Neurocognitive approach is the only systematic way of moving closer to the human level of competence in the linguistic area.

What should be done to further develop this approach?

- Analysis of experimental psychology results related to priming and connecting it to models of spreading activation.
- Analysis of neurolinguistic data, neuroimaging experiments related to word representation.
- Analysis of connectome project results, with evaluation of functions each area is involved in, to define the basis for brain-based representations, estimation of word valence and may other dimensions.
- Defining models that approximate neural activity by probability distributions and connecting them to vector models and statistical techniques of machine learning.
- Defining models for activation and inhibition spreading in networks connecting nodes that store vector representations of concepts.
- Developing algorithms to discover pathways of activations in human associative thinking during text comprehension, that will define weights in vector representation of concepts.
- Developing context-dependent concept representations on a large scale, adding such representations to each word sense in Wordnet.
- Creating and analyzing networks that represent episodes, model people, or other complex entities, by connecting concepts in self-consistent and language-independent way (creating graphs of consistent concepts).
- Collecting real-world information for semantic memory through analysis of ontologies, dictionaries, encyclopedia, Wikipedia, active search, visual editable representations, dialogues, computer games and collaborative projects.
- Creating models for different types of memory – recognition, semantic and episodic – and their interactions.
- Testing concept and acronym disambiguation algorithms that result from such modeling.
- Comparison with spectral methods, all forms of latent indexing, mapping, factorization, Bayesian and causal networks.
- Testing algorithms for creation and recognition of neologisms and made-up words that are fairly common in language.
- Applications or resulting algorithms in machine translation.
- Other applications: word games, semantic search, object recognition, interactive interviews, chat-based user modeling, semantic links between encyclopedia articles and other documents, automatic collection of biography, social networks and medical applications.

### 10. Psychiatric avatar scenario

Initial contact with psychiatrist takes relatively long time and may lead to misunderstandings and patient's suspicion if the doctor does not understand him/her immediately. Some patients do not mention their hallucinations and delusions treating them as less important than emotional problems they experi-ence. An interview with an avatar that collects information, has semantic memory and models of psychiatric diseases, thus asking relevant question leading to a summary and initial description of the problem, may not only save time, but may be easier for some patients that are introverted and do not want to open themselves up to other people, including psychiatrists.

The application will be based on semantic memory and the 20-question game algorithm. It will also collect a free text from the patient, either in typed or spoken form. For those patients that prefer to speak sound recording may provide additional source of useful information via the voice-stress analysis techniques that can discover hidden emotions. As a preliminary step in this direction we have analyzed 6 decision trees given in the DSM IV (Diagnostic and Statistical Manual of Mental Disorders) that allow for diagnosis in specialized psychiatric domains [127]. Asking relevant questions speeds up diagnosis, but in real application the approach should be different. Instead of collecting scripts and scenarios we should rather collect knowledge from textbooks and other sources that describe psychiatric diseases in terms of symptoms, complaints, features, medications, expert's opinions, and use these descriptions to form semantic memory. The information collected in this way allows for understanding patient's answers, maintaining a meaningful dialog, and narrowing down possible questions. The psychiatrists receives a digest that should help him to understand the problem better and facilitate establishing good contact with the patient.

Thus the psychiatric avatar will include the following elements:

1. Talking head (preliminary experiments based on Haptek head have been done).
2. Semantic memory containing structured knowledge
3. Scripts for initiation of the dialogue, ending, calling help and several other situations.
4. Control software based on the ideas taken from the 20-question game application.
5. Summarization of results.
6. Machine translation to make it multilingual.

A multilingual approach to collection of information may be very helpful at police stations. It frequently happens that communication with the police is difficult or impossible.

Another version of similar algorithm will be used in an Alzheimer Clinic (Bad Aibling, Bayern, Germany) to solicit personal stories from patients. There is good evidence that talking about family matters will strengthen the personal identity of patients and therefore preserves family contacts, improving quality of life and shorting the cognitive decline period. Talking heads that should be able to keep conversation and actively solicit information from people may play an important therapeutic role.

## 11. Conclusions and wider implications

General inspirations for neurocognitive linguistics have been outlined, drawing on recent experiments with priming, insight and creativity. This analysis has led to a proposal of a novel role for the non-dominant brain hemisphere: generalization over similar concepts and their relations, creation of abstract categories and complex relations among concepts that have no name, but are useful in reasoning and understanding. This activity has been partially captured in a new construct called *coset* defined for a linguistic concept as a set of concepts (partially overlapping with concepts that belong to other cosets) resulting from different relations of a given concept to other concepts. Such cosets, abstract concepts that lack phonological representation, have been used to enhance representation of texts. Approximations to the process of spreading brain activations based on probability waves is too difficult to use directly in text analysis, therefore "snapshots" of this activity has been identified with vector-based concept representation commonly used in statistical approach to NLP [66]. However, vector representation of concepts should not be reduced to statistical correlations with other concepts, as commonly done in text analysis. Snapshot of dynamic activity patterns, defining connections with other concepts, should also include structural properties of concepts, properties that rarely appear as contexts in typical texts, but may appear in textbooks presenting domain knowledge.

Relations between spreading activation networks, semantic networks and vector models have not yet been analyzed in details. The use of background knowledge in natural language processing is an important topic that may be approached from different perspectives. Without such knowledge analysis of texts, especially texts in technical or biomedical domains, is almost impossible. Neurolinguistic inspirations may be quite fruitful, leading to approximations of processes that are responsible for text understanding in human brains, but creation of useful numerical representation of various concepts is certainly a challenge. Large-scale semantic networks and spreading activation models may be constructed starting from large ontologies. For medical applications vector representation of concepts may be created by expansion of each term that is replaced by a coset using relationships provided by UMLS ontologies. To end up with a useful representation the utility of each new relation has to be checked, or a whole class of concepts based on some specific relations may be added to the coset and then pruned to remove concepts that are not useful in text interpretation. Association rules may also be helpful here [132]. A crude version of such approach has been presented here and already using the second-order expansions gave quite good results on a very difficult problem of summary discharge categorization. This is one of the approaches to enhance UMLS ontologies by vectors that represent these concepts in numerical way and could be used in variety of tasks. As far as we know this is the first practical algorithm that allows for a large-scale recreation of pathways of spreading brain activations in the head of an expert who reads the text.

Finding optimal enhanced feature spaces and simplest decompositions of medical records into classes using either sets of logical rules or minimum number of prototypes in the enhanced space is an interesting challenge. "Optimal" may here depend on a wider context as the meaning of a concept depends on the depth of knowledge an expert has. For example, family physician may understand some concepts in a different way than a cardiologist, but it should be possible to capture both perspectives using prototypes. A lot of knowledge that medical doctors gain through the years of practice is frequently never verbalized. Prototypes representing clusters of documents describing medical cases may be treated as a crude approximation to the activity of neural cell assemblies in the brain of a medical expert who thinks about a particular disease. This may be observed in clusterization of these documents if a proper space is defined. Clusters found in the documents containing hospital summary discharges may be interpreted in this way, although MDS mapping to two dimensions only has to introduce many distortions. It is relatively easy to collect information about rare cases that are subject to scientific investigation, but not the subtypes of the common ones. Finding such subclusters, or identifying different subtypes of disease, is an interesting goal that may potentially help to train young doctors by presenting optimal sets of cases for each specific cluster. It could also be potentially useful in more precise diagnoses. With sufficient amount of documents optimization of individual feature weights could also be attempted.

Although much remains to be done before unstructured medical documents and general web documents will be fully and reliably annotated in an automatic way, a priori knowledge certainly will be very important. Creating better approximations to the activity of the brain representing concepts and making inferences during sentence comprehension is a great challenge for neural modeling. In medical domain ontologies, relations between concepts, and classification of semantic types enables useful approximations to the neurolinguistic processes, while in general domains resources of this sort are still missing.